\documentclass[conference]{IEEEtran}
\IEEEoverridecommandlockouts
  \PassOptionsToPackage{numbers, compress}{natbib}

\usepackage{cite}
\usepackage{amsmath,amssymb,amsfonts}
\usepackage{algorithmic}
\usepackage{graphicx}
\usepackage{textcomp}
\usepackage{xcolor}
\usepackage[boxruled,linesnumbered]{algorithm2e}
\usepackage{graphicx} % graphics
\usepackage{hyperref}
\usepackage[british]{babel}
\usepackage{hhline}
\usepackage{multirow}
\DeclareMathOperator*{\argmax}{arg\,max}

\def\BibTeX{{\rm B\kern-.05em{\sc i\kern-.025em b}\kern-.08em
    T\kern-.1667em\lower.7ex\hbox{E}\kern-.125emX}}
\begin{document}

\title{Regional based query in graph active learning }

\author{\IEEEauthorblockN{1\textsuperscript{st} Roy Abel}
\IEEEauthorblockA{\textit{Department of Mathematics, Bar-Ilan University} \\
Ramat Gan , Israel \\
Email: royabel10@gmail.com}
\and
\IEEEauthorblockN{2\textsuperscript{nd} Yoram Louzoun}
\IEEEauthorblockA{\textit{Department of Mathematics, Bar-Ilan University} \\
Ramat Gan , Israel \\
Email: louzouy@math.biu.ac.il}
}

\maketitle

\begin{abstract}
Graph convolution networks (GCN) have emerged as the leading method to classify node classes in networks, and have reached the highest accuracy in multiple node classification tasks. In the absence of available tagged samples, active learning methods have been developed to obtain the highest accuracy using the minimal number of queries to an oracle. The current best active learning methods use the sample class uncertainty as selection criteria. However, in graph based classification, the class of each node is often related to the class of its neighbors. As such, the uncertainty in the class of a node's neighbor may be a more appropriate selection criterion.

We here propose two such criteria, one extending the classical uncertainty measure, and the other  extending the page-rank algorithm. We show that the latter is optimal when the fraction of tagged nodes is low, and when this fraction grows to one over the average degree, the regional uncertainty performs better than all existing methods. While we have tested this methods on graphs, such methods can be extended to any classification problem, where a distance metrics can be defined between the input samples.

All the code used can be accessed at : https://github.com/louzounlab/graph-al

All the datasets used can be accessed at : https://github.com/louzounlab/DataSets

\end{abstract}


\section{Introduction}
Relational information is often presented as graphs or multi-graphs, where nodes represent entities and edges represent relations between these entities.  Such relations can be used to predict the class of the nodes, using two main principles. The first and most used method is based on node class homophily, where neighboring nodes are assumed to belong to the same class with a high probability \cite{ji2012variance, berberidis2018data, zhu2003semi, zhu2003combining, sindhwani2005beyond, belkin2004semi}. This has been used in many propagation based algorithms where the class of a node is predicted using the class of neighboring nodes. The second approach presumes a correlation between the topological attributes (e.g. degree,centrality, clustering coefficient...) of nodes and their class \cite{shi2000normalized, yang2013community, rosen2015topological, naaman2018edge}. These two principles were combined in Graph Convolutional Networks (GCN). Such networks received much interest following the works of Kipf and Welling \cite{kipf2016semi}, where they have been shown to produce better accuracies than existing label propagation methods.  The main formalism proposed in such networks is the weighted combination of the input from previous layers:
\begin{equation}
X_{k+1}=\sigma(A*X_k*W_k),
\end{equation}
with  $W_K$ being the weights of the $k$ layer,  $X_k$ the input to this layer and $A$ a matrix derived from the adjacency matrix (e.g. a symmetric
 adjacency matrix with the identity matrix added to it). $X_0$ is usually external information about the nodes, but other methods including using the identity matrix \cite{schlichtkrull2018modeling}, nodes topological features or the frequency of neighbors belonging to each class in the training set \cite{benami2019topological} have been proposed.  The output of the network is then compared to known tags to compute the weights optimizing the fit.  Current GCN based methods presume a predefined set of such tags. However, often, such tags are expensive to obtain, and an active learning is used to query for the tags that would produce the highest precision (as defined through the prediction accuracy or any other measure) using the minimal number of tagged samples. Many Active Learning (AL) have been proposed \cite{settles2009active, lewis1994heterogeneous, culotta2005reducing,settles2008analysis}, most such methods are generic and do not use the graph information. We here show that combining the graph within the AL leads to significantly higher accuracies than simply applying generic AL to GCN.

\section{Related Work}

Graphs have been extensively used for machine learning, especially in  the context of GCN \cite{kipf2016semi, schlichtkrull2018modeling, berg2017graph} and GNN \cite{grover2016node2vec, rosen2015topological}. GCN were also used in combination with AL. However, in most such models,  the graph is only used for the ML part, while the AL is only performed on the nodes ignoring the graph structure. The most frequently used approaches in this domain are uncertainty sampling, and representative sampling \cite{settles2009active}.

Uncertainty sampling is a general framework for measuring informativeness \cite{lewis1994heterogeneous}, where a learner queries the instance that it is most uncertain how to label. 
Culotta and McCallum (2005) \cite{culotta2005reducing} employ a simple uncertainty-based strategy for sequence models called least confidence (LC): $\phi^{LC} (x)= 1- P(y^{*} |x; \Theta)$. Here, $y^{*}$ is the most likely label. This approach queries the instance for which the current model has the least confidence in its most likely labeling. 
Scheffer et al. (2001) \cite{scheffer2001active} propose another uncertainty strategy, which queries the instance with the smallest margin between the posteriors for its two most likely labels: $\phi^{M} (x) = P(y_1^{*}  |x; \Theta) - P(y_2^{*}  |x; \Theta)$, where $y_1^{*}$ and $y_2^{*}$ are the first and second best labels, respectively. Another uncertainty-based measure of informativeness is entropy (Shannon, 1948) \cite{shannon1948mathematical}. For a discrete random variable Y , the entropy is given by $H(Y )=- \sum(P(y_i) log(P(y_i)))$, and represents the information needed to "encode" the distribution of outcomes for Y . As such, it is often thought of as a measure of uncertainty in machine learning. 
We thus can calculate one of the uncertainty measures for each vertex based on the current model and choose the most uncertain instances to reveal.

In representative sampling, one assumes that  informative instances are  ``representative" of the underlying distribution. 
Fujii et al. \cite{fujii1998selective} considered a query strategy for nearest-neighbor methods that selects queries that are (i) least similar to the labeled instances, and (ii) most similar to the unlabeled instances.
Nguyen and Smeulders (2004) \cite{nguyen2004active} proposed a density-based approach that first clusters instances and tries to avoid querying outliers by propagating label information to instances in the same cluster. 
Settles and Craven (2008) suggested a general density-weighting technique combining both uncertainty and representative. They query instances as follows:
$\argmax_{X}{\phi_A(X) \times (\dfrac{1}{U} \sum_{U} {sim(X, X_U)})^\beta} $ where $\phi_A(X)$ represents the informativeness of x according to some ``base" query strategy A, and $U$ are the unlabeled samples. The second term weights the informativeness of x by its average similarity to all other instances in the input distribution (as approximated by U), subject to a parameter $\beta$ that controls the relative importance of the density term \cite{settles2008analysis}. 
Zhu and Wang also proposed sampling by a combination of uncertainty and density to solve the outliers problem emerging in some uncertainty techniques \cite{zhu2009active}

Beyond generic AL learning, recently a few works have proposed to use the graphs themselves for AL. Three main methods have been proposed: modularity, centrality and label propagation.
\begin{itemize}

\item In modularity based AL approaches, nodes are divided into communities. Macskassy proposed to reveal the one most central node in each community. Then divide each community into sub-communities and sample the most central node in each sub-community and so on. Since community based methods does not seem to work by themselves, Mackassy suggested a hybrid method combining communities, centrality. and uncertainty with Empirical Risk Minimization (ERM) framework \cite{macskassy2009using}.
Ping and Zhu proposed combining communities structure to perform batch mode active learning. They used communities to consider the overlap in information content among the "best" instances \cite{ping2017batch}.

\item Centrality based approaches focus on nodes which are more central (e.g. higher degree). The assumption is that the  central nodes will have a major impact on the unknown labels, as used by Macskassy  in the  ERM algorithm, where he showed that betweenness centrality is a good measure for centrality \cite{macskassy2009using}.
Cai and Chang proposed  to calculate a node representativeness score based on graph centrality. They tested several centrality measures: degree centrality, betweenness centrality, harmonic centrality, closeness centrality, and page-rank centrality. They conclude that the Page-Rank centrality is superior, and suggest using it when the prediction model is not informative enough \cite{cai2017active}. 

\item In label propagation approaches, the implicit assumption is of label  smoothness over the graph or over the projection of the graph into some 
manifold in $R^{N}$. Ming Ji proposed to select the data points to label such that the total variance of the Gaussian field over unlabeled examples, as well as the expected prediction error of the harmonic Gaussian field classifier, is minimized. An efficient computation scheme was then proposed to solve the corresponding optimization problem with no additional parameter \cite{ji2012variance}. Yifei Ma extended sub-modularity guarantees from V-optimality to $\Sigma$ -optimality using properties specific to Gaussian Markov Random Field  (GRMF)s  \cite{ma2013sigma}.
Dimitris Berberidis proposed to sample the nodes with the highest expected change of the unknown labels. Thus, in contrast with the expected error reduction and entropy minimization approaches that actively sample with the goal of increasing the “confidence” on the model, Berberidis et al. focus on maximally perturbing  the model with each node sampled. The intuition behind this approach is that by sampling nodes with the largest impact, one may take faster steps towards an increasingly accurate model \cite{berberidis2018data}.
\end{itemize}

We here propose an alternative approach, which is to detect regions of 
uncertainty in the graph and sample those regions (in contrast with uncertain nodes). While the sampling of nodes with high uncertainty was proposed in a vast array of applications \cite{cai2017active, zhou2014active, macskassy2009using}, sampling using regional uncertainty (and in the cases of directed graphs, regional directed uncertainty) has never been successfully applied.  Chen et al. recently proposed regional uncertainty algorithms, but obtained very low accuracy \cite{chen2019activehne}.

\section{Main contributions of the current work}
The current work proposes two novel claims, and a suggested mechanism:
\begin{itemize}
\item We propose to replace the node entropy by the regional entropy leads to a faster increase in accuracy in GCN based AL methods.
\item We propose an extension of the PageRank algorithm to detect nodes shedding information on  unlabeled nodes in directed graphs.  
\item {\it We propose that information can be gained from the graph, only when the sampling rate is low enough so that most unlabeled nodes have no labeled networks.}
\end{itemize}
Note that the last item is only a suggestion and we do not bring a clear proof for it, only a comparison between similarity and the outcome of the AL.
\section{Model and data}
\subsection{GCN in directed graphs.}
Given a graph $G=(V,E)$, with an adjacency matrix $A\in R^{N \times N}$ (binary or weighted), a diagonal degree matrix D , a node feature matrix $X_0\in R^{N \times F}$ (i.e., F-dimensional feature vector for N nodes), and a label matrix for labelled nodes Y. A multi-layer Graph Convolutional Network (GCN) is a deep learning model, whose layers are defined as Eq. 1. The last layer is a soft-max used to determine the probability of each category given the input.  The adjacency matrix is typically replaced by a normalized and symmetrized version (e.g. $D^{-1/2} (A+A^T+1) D^{-1/2})$. We  use an extension to this model by incorporation of an asymmetric adjacency matrix \cite{benami2019topological}. In this model we do not lose the direction information in  the graph. We incorporate the direction by separating  the adjacency matrix (asymmetric in directed graph) into its symmetric and anti-symmetric components and concatenate them – creating a $2n \times n$ adjacency matrix. The dimension of the output of each layer is: $[(2N \times N) \times (N \times i_n ) \times (i_n \times o_n )]=2N \times o_n$, which in turn is passed to the next layer following a rearrangement of the output by splitting and concatenating it to change dimensions from  - $2N \times O_n$  to $N \times 2O_n$. 

\subsection{Active Learning}

We have developed several graph-oriented and used several non-graph oriented methods to choose the next instance to reveal.

 In all methods,  we use a greedy framework where the nodes with the highest score based on  the network structure and knowledge gained so far are selected. The only difference between the methods is the score used.  At each iteration, batch of nodes are queried for their labels. Ideally, we would like to query one vertex at a time, but there is a trade-off between the learning success and the running time. Unless explicitly stated, the batch size is 1.

The following classical AL methods have been tested:
\begin{itemize}
\item Centrality  implemented using the  Page Rank algorithm.
\item Entropy - Shannon entropy.			
\item Geo Dist - We Calculate shortest path's length of all pairs of known and unknown nodes.	The score is the minimum distance from a certain node to all known nodes.If there is no path connecting a node to any known node, a (high) score of 9 is given.
\item	 Geo Centrality - combination of geodesic distance and centrality. The two scores are normalized and the merged score is $0.7 * geo dist + 0.3 * centrality$		
\item Margin - the absolute difference between the probabilities of the two most likely classes.
\item Random - choose a node at random.			
\item Rep Dist - Using the last layer output of the GCN model as a $R^C$ representation vector.	Then choosing nodes which are different from labeled nodes to get representative distribution. We checked two distance definitions: MAH -  Mahalanobis distance. LOF -  Local Outlier Factor provided by "sklearn" package.	
\item Feature AB - Attractor Basin \cite{muchnik2007self} 			
\item K Truss - Extension to k-core. $T_k$ is the largest subgraph such that all edges belong to at least $K-2$ triangles. 	Node $v_i$ has a K-truss score $k$ if it belongs to the $k$ K-truss subgraph, but not to any $k+1$ K-truss subgraph. Nodes with high K-Truss are argued to have a high influence on other nodes  \cite{malliaros2016locating}.		
\item Chang Based - This method is inspired by Cai and Chang \cite{cai2017active}.	 It is a combination of centrality, uncertainty, and density.		
	According to their experiments the best combination is starting with high centrality, and later raise the uncertainty and density.		
	Similar to Cai and Chang, uncertainty is entropy and centrality Page Rank, but instead of density we took the representativeness measure of rep dis mahalanobis.		
	
\end{itemize}
\subsection{Regional Active Learning}
As mentioned before, in many cases labels are correlated within regions in the graph. Thus the uncertainty of the entire region can provide more information than the uncertainty of the node itself. Furthermore, observing the entire region can reduce outliers queries (e.g. nodes disconnected from the network), which is one of the main downsides of uncertainty sampling. 

Specifically, we calculate for each node the probability of all classes. Then for each node $v_i$ the regional probability is the average of the probabilities over the region 
\begin{equation}
\tilde{p}(y_i = c) = \frac{1}{|N_i|}\sum_{v_j \in N_i} P(y_j = c)
\end{equation}
where $N_i$ is the region (currently defined as first neighbors of the node $i$ in the undirected graph underlying the directed graph, but can be defined to be more distant or directed regions)  and $y_i$ is the label of node i.
Regions where all nodes have a high predicted probability to belong the same class will have high regional probability for this class, while regions with high variety will have low regional probability for all classes (Fig. 1). This regional probability can be used as input to any uncertainty measure.

An alternative approach is the computation of  the uncertainty measure for each node separately. In such a case, the regional uncertainty is calculated by averaging the score over the region: 
\begin{equation}
\phi (v_i) = \frac{1}{|N_i|}\sum_{v_j \in N_i} \phi (v_j)
\end{equation}
where $\phi (v_i)$ is the uncertainty score of node i.
This regional measure would give a high uncertainty to nodes close to many uncertain nodes (and not to nodes in heterogeneous regions), thus reducing uncertainty over wide areas speeding up the active learning process. 

Both  measures scale easily to large graphs, since calculating each node's region is required only once.

\begin{figure}
 \includegraphics[width=12cm, height=18cm]{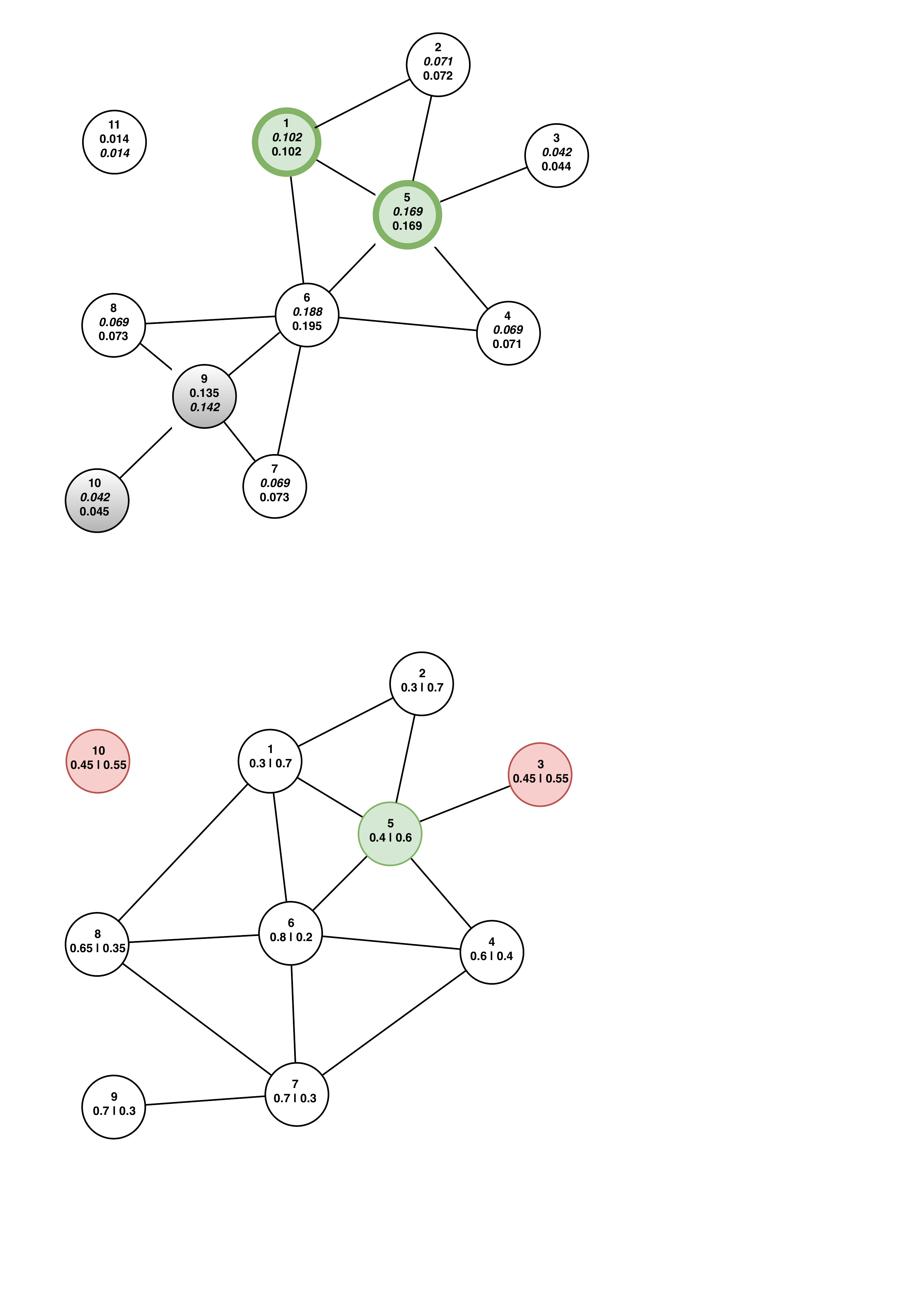}
\caption{\it{Schematic figures: Lower plot - Regional uncertainty: Given the classes' probability of each node marked as $P(y = class 1) | P(y = class 2)$ (resulting from the classification or from a prior), local uncertainty techniques would give the highest score to nodes 3 and 10 (marked in red). However, knowing the label of these nodes would have a minimal effect on their region. We wish to reveal the label of the node with the most uncertain region. Therefore, we average the probabilities over the neighbors.$ \tilde{p}$ of node 5 will be (0.49, 0.51), thus node 5 (marked in green) will be the node with the most uncertain region. Nodes without neighbors have a regional uncertainty of zero per definition. Upper plot - Adaptive Page Rank (APR). We assume that nodes 1 and 5 (marked in green with thick borders) are already labeled. For each node, the PR and APR are give (APR is in italics). For labeled nodes or nodes not connected to labeled nodes, the APR and PR are equal.  Standard centrality measures would choose node 6 (since it is most central), ignoring the fact that this node is close to the two already known nodes, and thus probably easy to predict. The ratio between APR and PR would favor nodes 9, 10 (marked in grey), since those are far from the influence of nodes with known labels and will append more new information.}}
\end{figure}

\subsection{Adaptive Page Rank}
While many use Page Rank as a measure of centrality to choose informative nodes to query, Page Rank is based only on the graph structure ignoring information already obtained from labeled nodes. We here propose an adaptive extension to page rank which also considers sub-graph of known nodes. As will be shown, this method outperforms using the standard page rank since it is adaptive to the nodes already queried, hence more suitable for active learning tasks. This method is useful especially in cases where part of the graph nodes' labels is already known.  
Page Rank \cite{page1999pagerank} can be thought as diffusion or a random walk process where each node receives an initial rank of $\frac{1}{N}$, and at each iteration passes most of its rank through its out edges evenly. The rest is passed evenly to all nodes in the graph.
The Adaptive Page Rank (APR) follows the same concept, but the initial rank of each node is the standard page rank if the label is known and 0 otherwise. Then we start a random process where for each node, a random node is chosen with a probability of gamma, and one of the nodes' neighbors is chosen with a probabiltiy of $1-\gamma$, and the rank is passed to the chosen node.  This process is repeated until convergence is obtained. The Rank of known nodes is fixed to be the regular page rank across all iterations. 

Calculating the steady state of APR (the vector of APR values over all nodes) can be done by solving the equation:
\begin{equation}
APR = \gamma(\overline{A}^T \times APR) + (1-\gamma)(\frac{1}{N})
\end{equation}
where $N$ is the number of nodes, APR is a $R^N$ vector of the scores, $\gamma \in (0,1)$, and $\overline{A} = I - L^{RW}$  where $L^{RW} = I - D^{-1} A$ is the random-walk normalized Laplacian matrix, A is the adjacency matrix and D is the degree matrix. 
Since the values of APR are fixed for the labeled nodes, we can solve the equation above through:
\begin{equation}
APR(U)=  \gamma \overline{A}_{U:L}^{T} \times APR(K) + \gamma\overline{A}_{U:U}^{T}\times APR(U)  +\frac{1- \gamma}{N},
\end{equation}
where U represents the indices of the unknown nodes and L represents the labeled nodes, leading to:
\begin{equation}
APR(U)=  (I- \gamma\overline{A}_{U:U}^{T})^{-1} (\gamma \overline{A}_{U:L}^{T}\times APR(L)+ \frac{1-\gamma}{N})
\end{equation}
The adaptive page rank received can be described as the influence of the labeled nodes over the unknown. We thus can use this measure in our active learning framework seeking the best ration between APR and PR to choose nodes which on the one hand are not affected by the known nodes, and on the other hand would affect greatly on the other nodes. 

\subsection{Data Sets}
The names, edge number, node number and number of different classes in each dataset are detailed in table 1.

\subsubsection{Cora}
The Cora dataset \cite{mccallum2000automating} consists of 2,708 scientific machine learning publications categorized into one of seven topics. The citation network consists of 5,429 links. We consider only papers which are cited by or cite other papers. 
Bag of words is also available for each node. Each publication in the dataset can be described by a 0/1-valued word vector indicating the absence/presence of the corresponding word from the dictionary. After stemming and removing stop words and rare words (frequency less than 10), the dictionary contains 1,433 unique words \cite{sen2008collective}.

\subsubsection{CiteSeer for Document Classification}
The CiteSeer data set \cite{giles1998citeseer} consists of 3,312 scientific publications classified into six categories: Agents, Artificial Intelligence, Database, Human Computer Interaction, Machine Learning and Information Retrieval. There are 4,732 links describing citations in the data set. 
Bag of words is also available for each node. Each publication in the dataset can be described by a 0/1-valued word vector indicating the absence/presence of the corresponding word from the dictionary. The dictionary consists of 3,703 unique words after stemming and removing stop words and rare words (frequency less than 10) \cite{sen2008collective}.

\subsubsection{Email-Eu-core network}
The Email-Eu-core network \cite{leskovec2007graph, snapnets, yin2017local} was generated using email data from a large European research institution. There are 25,571 edges (u,v) in the network describing a person $u$ sent person $v$ at least one email. The emails only represent communication between institution members, core of 1,005 people, and the dataset does not contain incoming messages from or outgoing messages to the rest of the world.
The dataset also contains "ground-truth" community memberships of the nodes. Each individual belongs to exactly one of 42 departments at the research institute.
This network represents the "core" of the email-EuAll network, which also contains links between members of the institution and people outside of the institution (although the node IDs are not the same).

\subsubsection{PubMed Diabetes}
The PubMed Diabetes dataset consists of 19,717 scientific publications from the PubMed database pertaining to diabetes classified into one of three classes. The citation network consists of 44,338 links. 
Bag of words is also available, each publication in the dataset is described by a TF/IDF weighted word vector from a dictionary which consists of 500 unique words \cite{sen2008collective}. 

\subsubsection{Subelj Cora}
Citation network of 23,166 scientific computer science publications classified into one of ten categories: Artificial Intelligence, Operating Systems, Data Structures Algorithms and Theory, Programming, Networking, Encryption and Compression, Human Computer Interaction, Databases, Hardware and Architecture, Information Retrieval. The citation network consists of 91,500 links indicating that the left node cited the right node \cite{konect:2017:subelj_cora}.

\subsubsection{Wikispeedia navigation paths}
This dataset is collected from the human-computation game Wikispeedia. In Wikispeedia, users are asked to navigate from a given source to a given target article, by only clicking Wikipedia links. A condensed version of Wikipedia is used, with 4,604 articles, and 119,882 directed links connecting them.
Each article is classified by its subject to one of the following: History, People, Countries, Geography, Business Studies, Science, Everyday life, Design and Technology, Music, IT, Language and Literature, Mathematics, Religion, Art, Citizenship \cite{west2012human, west2009wikispeedia}.
\begin{table}[htbp]
\caption{}
\begin{center}
\begin{tabular}{|c|c|c|c|}
\hline
\textbf{Data Set} & \textbf{\textit{Nodes}}& \textbf{\textit{Edges}}& \textbf{\textit{Classes}} \\
\hline
CORA&	2,708&	5,429&	7\\
\hline
CITESEER&	3,312&	4,732&	6 \\
\hline
EMAIL-EU&	1,005&	25,571&	42 \\
\hline
PUBMED&	19,717&	44,338&	3 \\
\hline
SUBELJ CORA& 	23,166&	91,500&	10 \\
\hline
WIKISPEEDIA&	4,604&	119,882&	15 \\
\hline
\multicolumn{4}{l}{}
\end{tabular}
\label{tab1}
\end{center}
\end{table}
\section{Results}
\subsection{In all studied datasets, GCN with neighbors tags as input outperforms other learning approaches}
In order to test the effect of different AL schemes, we first tested the best machine learning framework for the inference task in three of the  datasets studied here (Table 1). We have tested four algorithms: GXBoost, FFN, GCN and RF. For each algorithm, we tested three types of input, the neighbors tag, topological features of the node and the combination of the two, as proposed by Benami et al. \cite{benami2019topological}. The precision was computed for different training set fractions in a passive setup where the training set is pre-defined. We have not performed any parameter tuning, or optimization of the machine learning, since out goal was to test the difference between AL schemes. In the vast majority of datasets studied, a GCN with the number of neighbors in the training set belonging to each class obtained the highest accuracy and Micro and Macro F1 (Fig 2). Thus, we only used this setup for all the AL schemes. The following parameters were used in the different algorithms. In the Random Forest, we used 100 estimators, and a balanced class weight.  In the XGBoost, we used 15\% internal validation, Dart boosters, a max depth of 7, $\lambda=1.3.\eta=1.3, \gamma=3$, a rate drop of 0.2, weighted sampling,  a softprob objective function, and early stopping after 10 steps. In the GCN ,  we used hidden layer sizes of 16, 200 epochs, a learning rate of 0.01, Relu non-linearities, a drop out of 0.6 a weight decay of 0.005  and an internal validation of 10\%.

\begin{figure}
 \includegraphics[width=10cm, height=8cm]{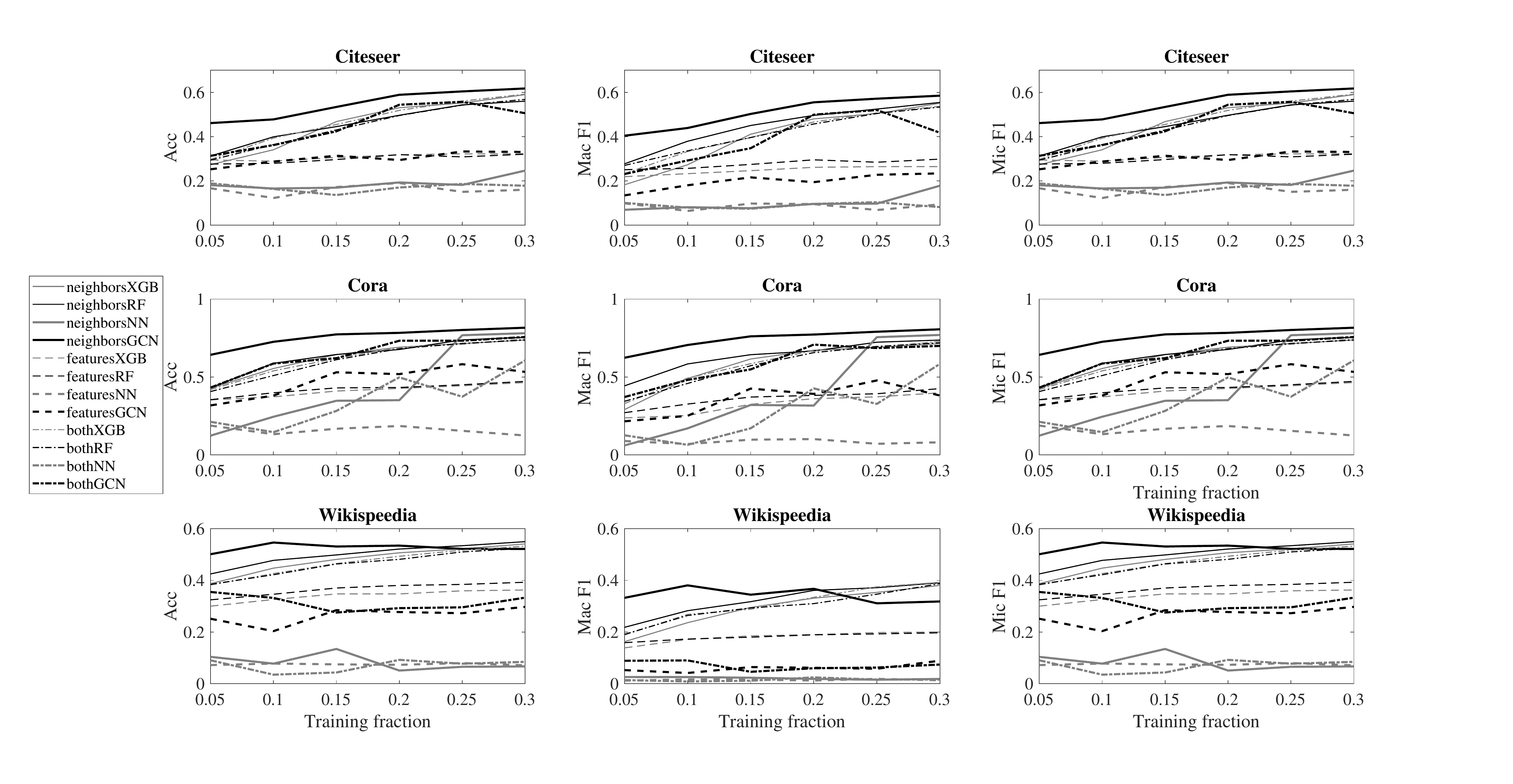}
\caption{\it{Accuracy, Micro and Macro F1 as a function of training set fraction in 3 datasets  for different learning methods. We tested for three reported datasets multiple precision estimates as a function of the training set fraction. We have tested four algorithms: GXBoost, FFN, GCN and RF. For each algorithm, we tested three types of input, the neighbors tag, topological features of the node and the combination of the two. The precision was computed for different training set fractions in a passive setup where the training set is pre-defined. One can clearly see that in all datasets and using all measures, the GCN with the neighbors tags as input produces the best accuracies. No parameter fitting was performed. Thus, in principle, different methods could produce higher precisions than shown here in any of the three measures used.}}
\end{figure}

\subsection{Among local methods entropy and geo-centrality  methods reach highest accuracy}
We first tested a large array of local AL approaches (as detailed in methods). For each dataset studied, we computed the loss of the model, the accuracy and micro and macro F1 of unlabeled nodes for the different AL approaches. The initial state was 1 random sample of each class, and the batch size used was 1, except for the Pubmed and Subelj datasets, where we used a batch size of 5. In the Cora and CiteSeer dataset,  bag of word information was also available. We thus tested in these two datasets either a topology based classification, or a content based classificaiton. In the latter, we stopped after 200 classified nodes, since the accuracy was already very high at this stage. When comparing the AL with random sampling, many of the AL scheme actually perform much worse than random (Fig 3 upper plot for accuracy as a function of the sampled fraction and lower plot for the difference between the loss (of GCN), accuracy and micro and macro F1 at the last time point and the one expected from a random sampling). In average the geo-centrality and entropy produce the best accuracy for different datasets for most sampling rates and for the last time point. No major differences were detected between the micro and macro F1 and between the accuracy. The F1 score and accuracies were computed using the default setting of the F1 score of sklearn. All simulations were repeated 20 times, we do not plot standard errors to avoid cluttering the figures, but the average standard error was less than 0.015, and is thus much smaller than the difference between typical methods. 
\begin{figure}
 \includegraphics[width=10cm, height=8cm]{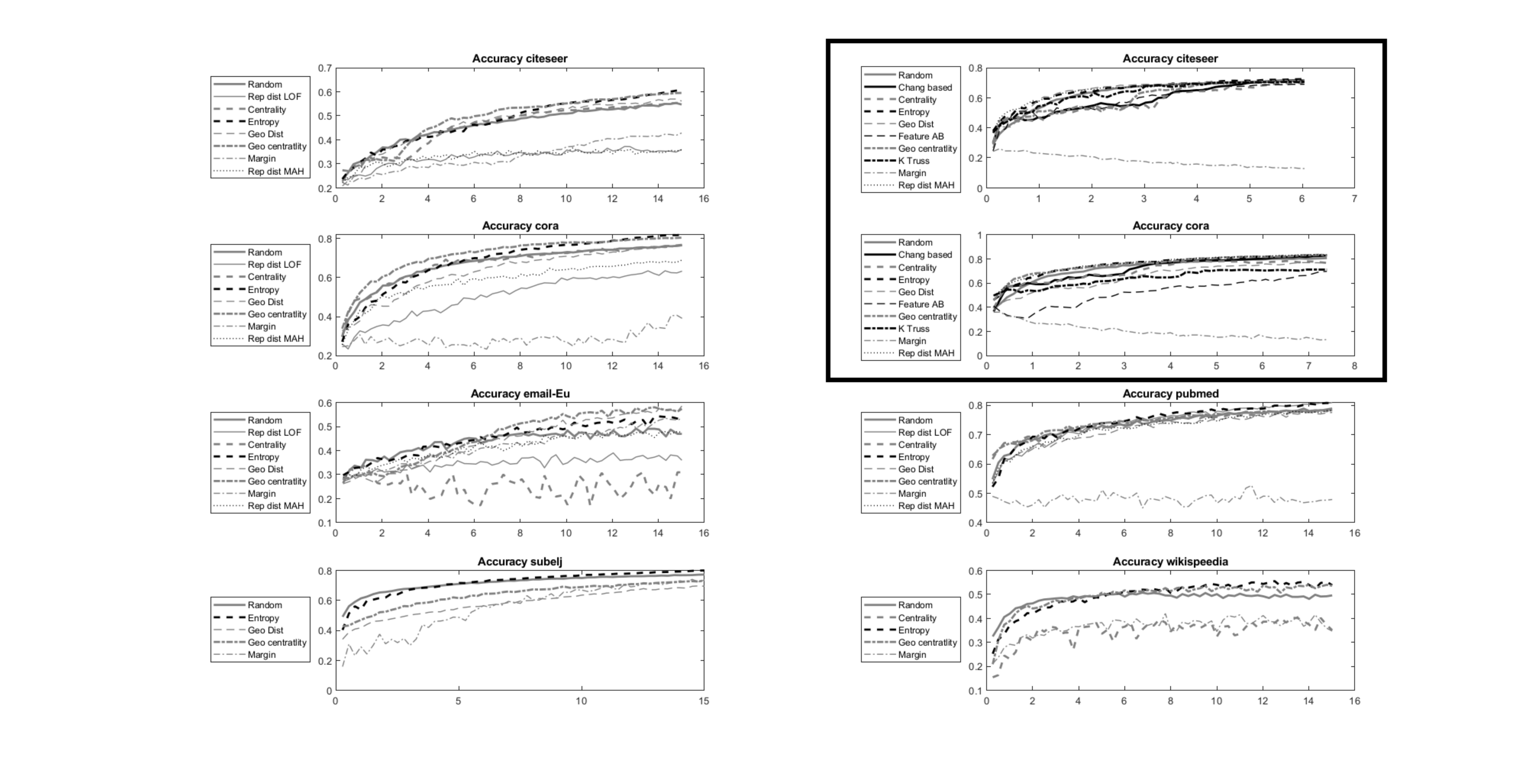} 
 \includegraphics[width=10cm, height=8cm]{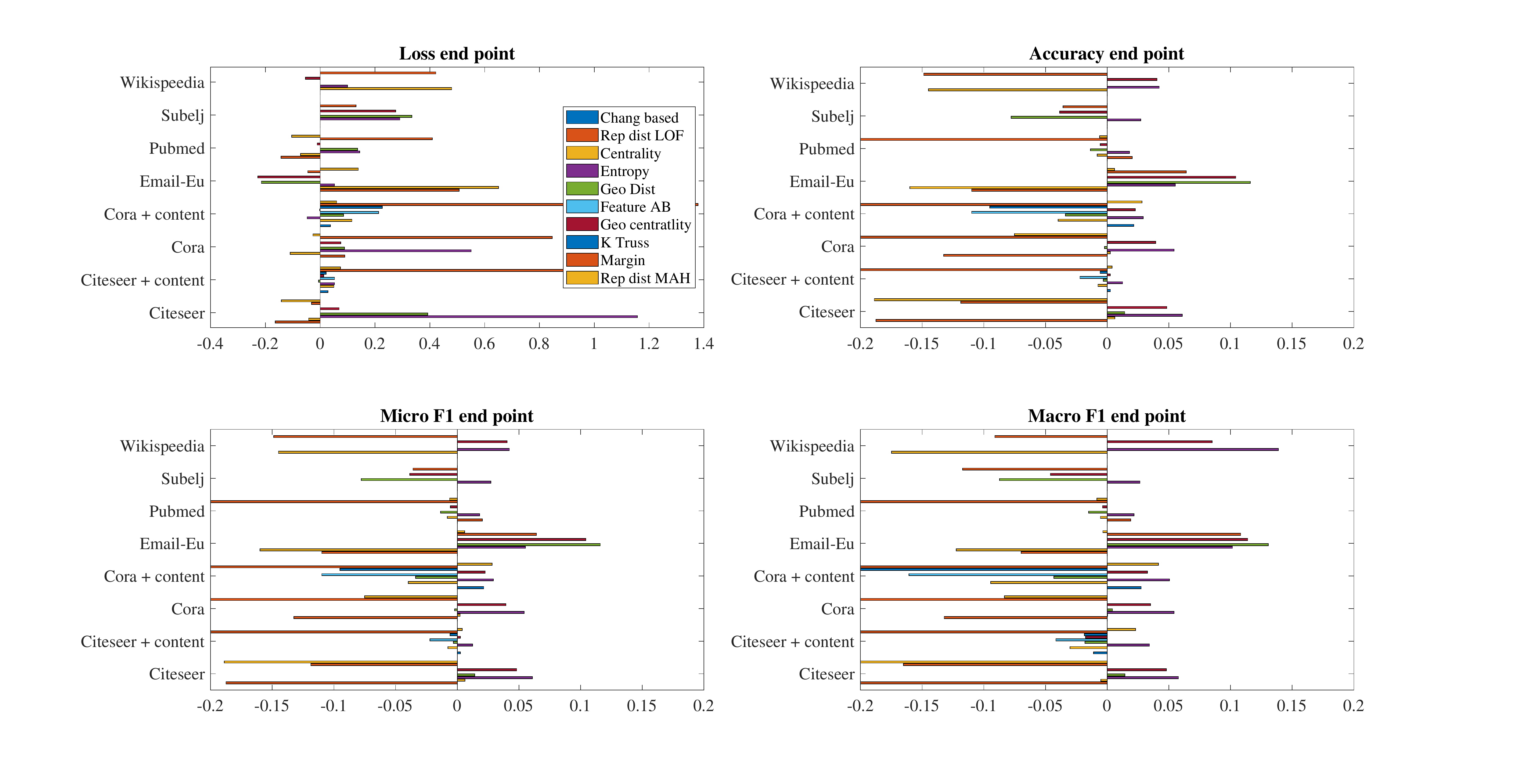}
\caption{\it{Upper plot. Accuracy as a function of sampling fraction for different datasets and different local AL methods.  results lower than the random sampling show AL algorithms that are worse than random sampling.  The two subplots surrounded by a box are the results with the BOW. Lower plot difference between random sampling and each of the AL methods at the last time point sampled (15 \% when no input is used and 200 nodes when external information is used). We compared four measures: The GCN loss function, the accuracy and the micro and macro F1 values.  Positive values for the accuracy and F1 or negative values for the loss show AL schemes that are better than random. Many AL schemes are actually worse than random, and as such should not be used for network based ML. }}
\end{figure}
\subsection{Regional methods are better than local methods and among regional, entropy and margin are best}
We tested the accuracy obtained using the regional AL scheme. The analysis was similar to the local AL schemes with the same number of simulations and setup. However, the results are very different. In Fig 4, we present the difference between the accuracy/F1/Loss obtained in the regional AL scheme and in the random sampling. Most values are positive for the accuracy and F1 scores, and negative for the loss, since a lower loss implies a higher precision. While in most datasets studied , most regional scheme produce a significant performance boost (paired t test at last time point of the average of accuracy in regional AL scheme vs random $p<0.05$ for all datasets, except for Wiki and Email-EU). For these last two datasets an interesting difference can be observed. We have tested two averaging scheme. Either first computing the distribution of class predicted probabilities over neighbors $\tilde{p}$ and only computing the score (the entropy or the margin between the probability of first and second most probable predictions) on the average probability or first computing the same score over all neighboring nodes and then averaging the score (denoted by AE in Fig 4). In the  Wiki and Email-EU dataset, the AE approach works very well, while averaging the probabilities of neighbors actually performs much worse than random. averaging the local uncertainties over the region actually drastically improves performance (positive values in the bottom plot of figure 4). To further test the accuracy that can be obtained using the regional AL, we compared the performance of the different algorithms tested here and compared them to existing performances in the standard Cora and CiteSeer datasets (Table II). For fair comparison we also fixed 1,000 nodes as test, and 500 nodes as validation set to limit the pool of nodes that can be queried by our algorithm, even though no validation is required at the proposed techniques. Note that each algorithm was tested 750 times (5 (test sets) $\times$ 10 (validation sets) $\times$ 15 (initial labelled sets)). While we outperform all existing methods in the Cora dataset, and get very similar results to the best accuracy obtained by Chang et al methods, with much simpler algorithms. 

\begin{figure}
 \includegraphics[width=10cm, height=8cm]{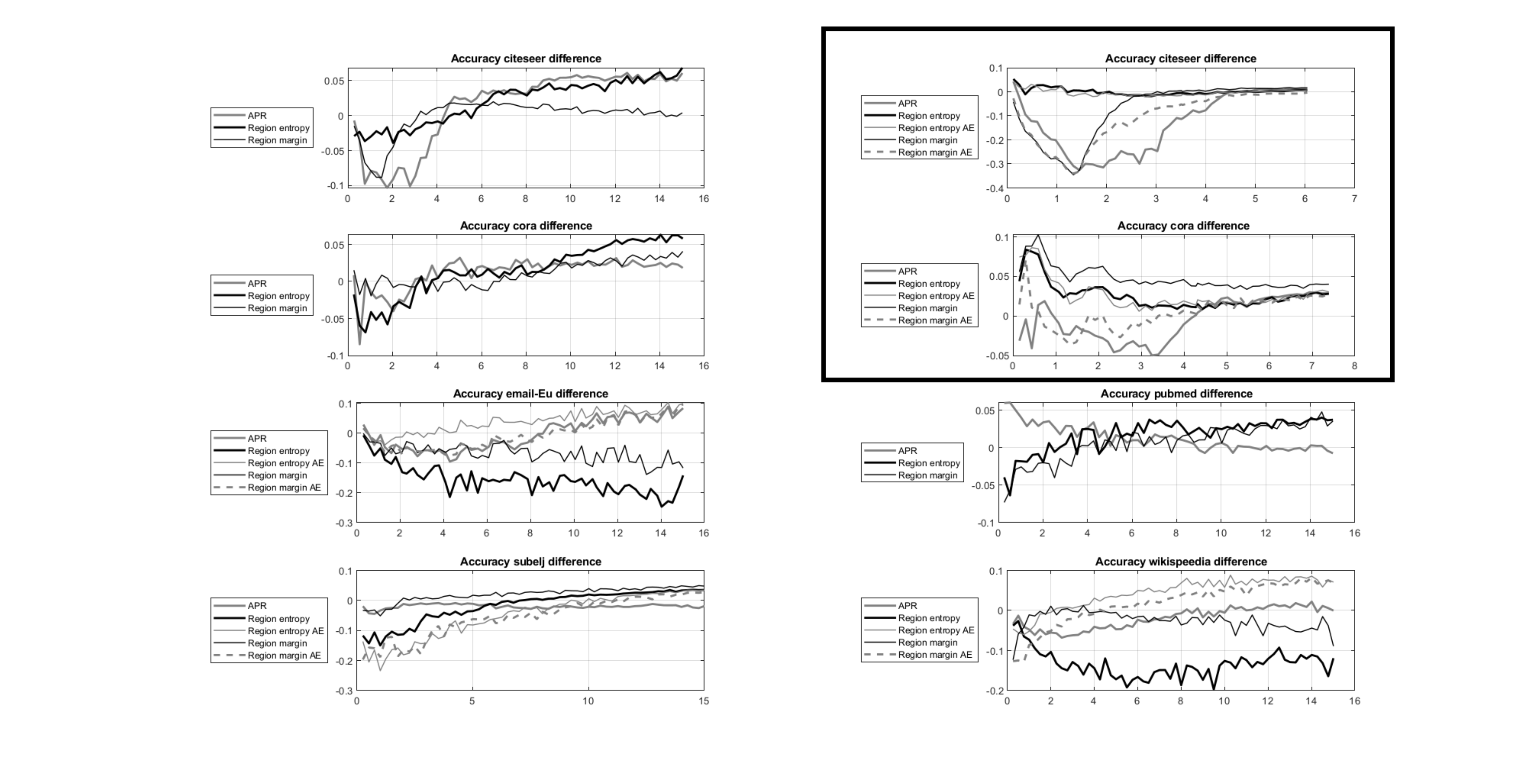} 
 \includegraphics[width=10cm, height=8cm]{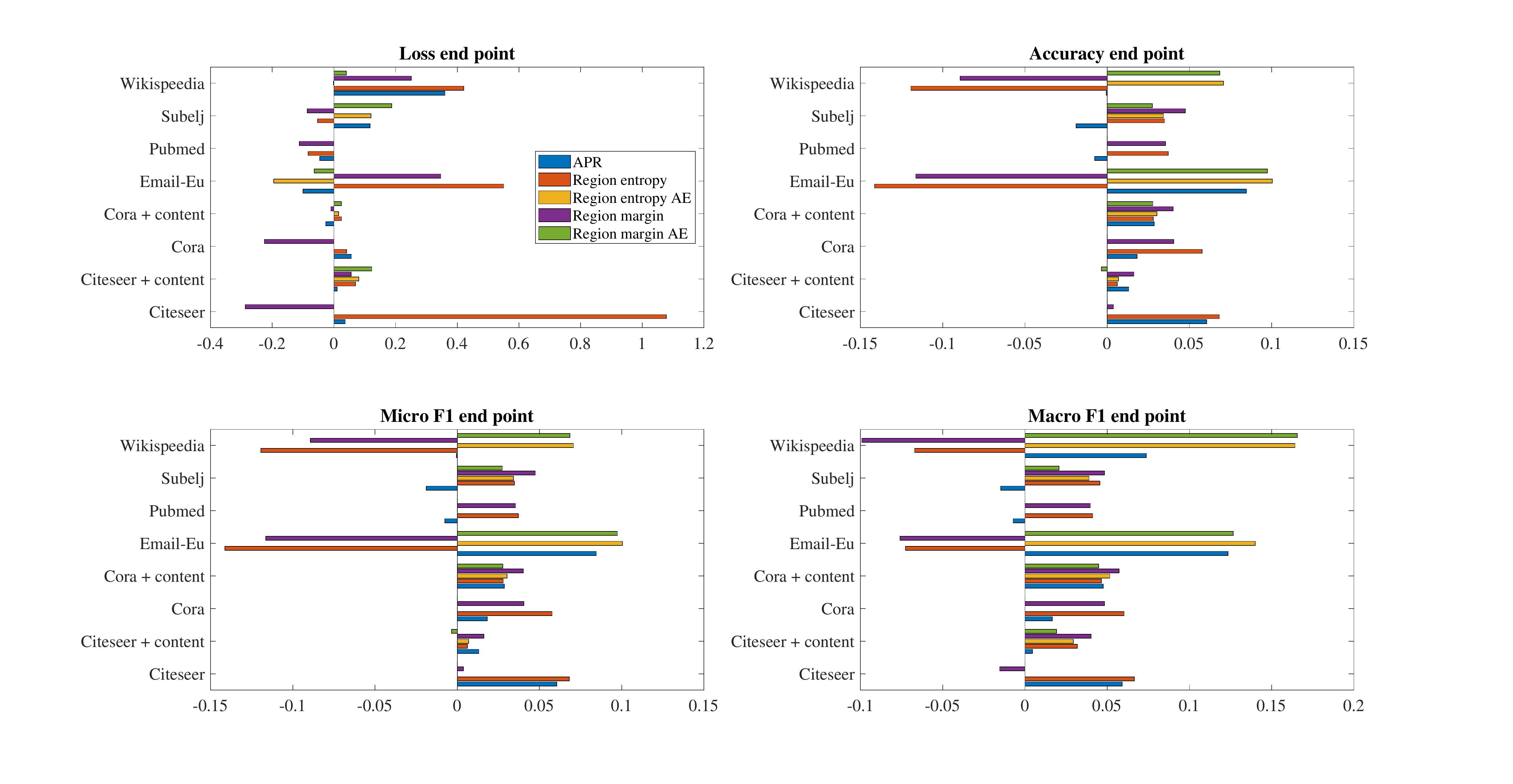}
\caption{\it{Upper plot and lower plot follow figure 3. The only difference is that the upper plot in this figure is the difference with random, so positive values represent higher accuracy than the one obtained in random sampling, while negative accuracy are worse than random sampling. }}
\end{figure}

\begin{table*}
  \centering
    \caption{Results}
  \renewcommand{\arraystretch}{1.2}
  \begin{tabular}{|p{3.5cm}|c|c|c|c|c|c|}
    \hline
    \multirow{2}{3.5cm}{\textbf{algorithm}} & \multicolumn{3}{c|}{\textbf{Cora}} & \multicolumn{3}{c|}{\textbf{CiteSeer}}\\
 
    \cline{2-7}
    & \textbf{accuracy} & \textbf{macro-F1} & \textbf{micro-F1} & \textbf{accuracy} & \textbf{macro-F1} & \textbf{micro-F1}\\
    \hline
    our random & 0.803 & 0.779 & 0.803 & 0.697 & 0.627 & 0.697 \\ \hline
    Kipf (random) \cite{kipf2016semi} & 0.801 & - & - & 0.679 & - & -   \\ \hline
    HNE \cite{chen2019activehne} & 0.59* & - & - & - & - & - \\ \hline
    TV/MSD** \cite{berberidis2018data} & 0.78* & - & - & 0.7* & - & - \\ \hline
    ClassSeek** \cite{mcdowell2015relational} & 0.805* & - & - & 0.695* & - & - \\ \hline
    $\Sigma$-Opt** \cite{ma2013sigma} & 0.73* & - & - & \textbf{0.71*} & - & -  \\ \hline
    ALFNET \cite{bilgic2010active} & 0.78* & - & - & 0.7* & - & -  \\ \hline
    Chang const param \cite{cai2017active} & - & 0.811 & 0.823 & - & 0.660 & 0.701 \\ \hline
    Chang adaptive param \cite{cai2017active} & - & 0.812 & 0.824 & - & \textbf{0.669} & \textbf{0.720} \\ \hline
    PR & 0.802 & 0.770 & 0.802 & 0.691 & 0.606 & 0.691 \\ \hline
    APR & 0.815 & 0.788 & 0.815 & 0.693 & 0.615 & 0.693 \\ \hline
    entropy & 0.803 & 0.785 & 0.803 & 0.696 & 0.628 & 0.696 \\ \hline
    region entropy & 0.814 & 0.804 & 0.814 & 0.691 & 0.645 & 0.691 \\ \hline
    margin & 0.800 & 0.782 & 0.800 & 0.700 & 0.631 & 0.700 \\ \hline
    region margin & \textbf{0.827} & \textbf{0.814} & \textbf{0.827} & 0.705 & 0.647 & 0.705 \\ \hline

\multicolumn{7}{l}{$^{\mathrm{*}}$ score is estimated from figure} \\
\multicolumn{7}{l}{$^{\mathrm{**}}$ scores for smaller budget, since it was the maximum tested in this work}

  \end{tabular}

\end{table*}
\subsection{The gain of regional methods occurs at  low sampling rates.}
An interesting result from the comparison of regional AL schemes is that APR outperforms all other methods at low sampling fractions (typically less than 5 \%). To understand, why  the regional based APR only helps at low sampling fractions, we computed the distance of a typical node to randomly sampled node, as a function of the sampling fraction.  This distance drops to a plateau around 5 \% sampling suggesting that beyond this fraction, sampling more nodes does not provide a significant regional advantage. We suggest this conjecture as a general method to estimate the sampling range where regional methods are advantageous and now plan to study it in detail.
\begin{figure}
 \includegraphics[width=8cm, height=6cm]{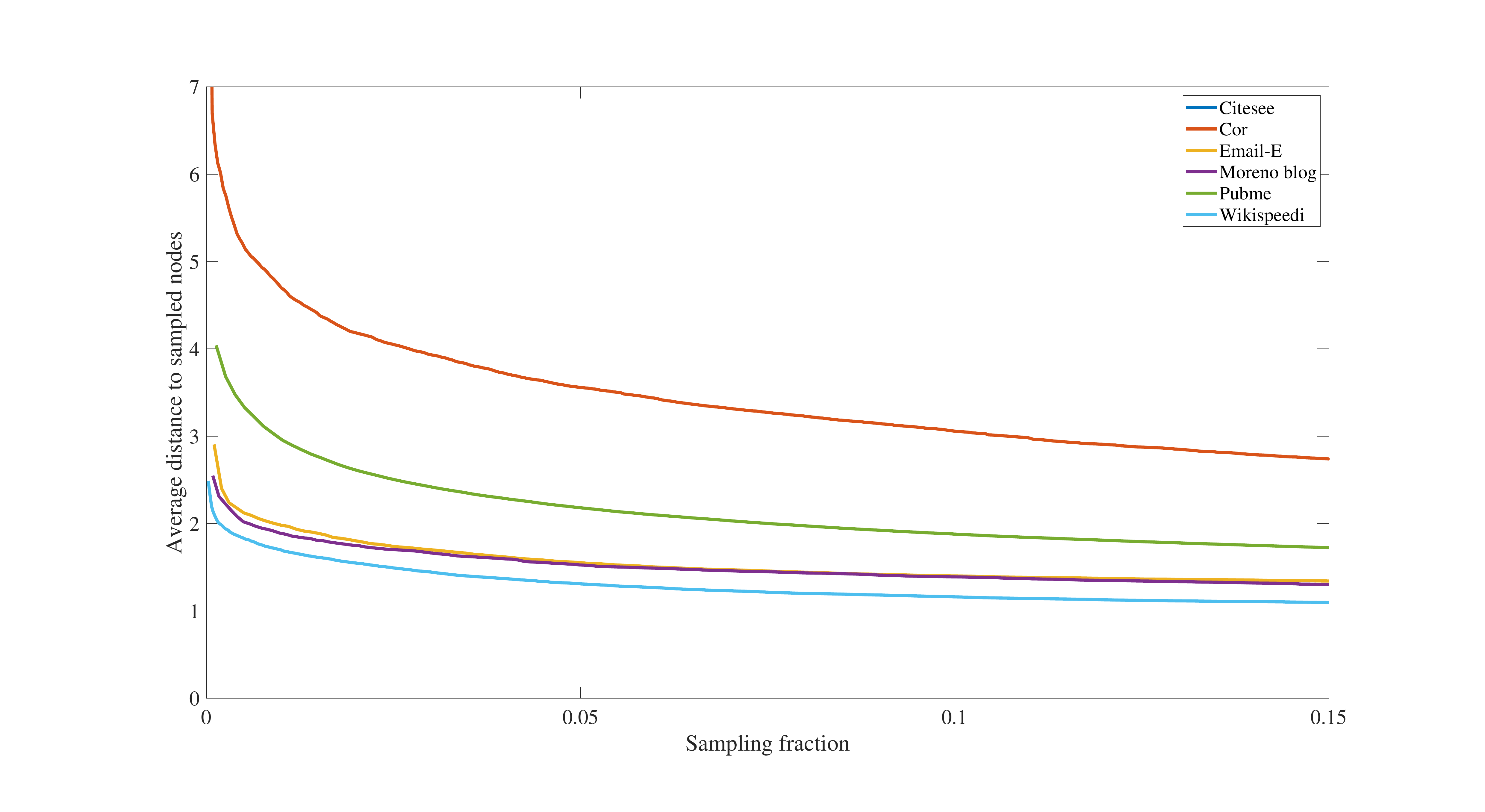} 
\caption{\it{Distance to randomly sampled nodes as a function of the fraction of sampled fraction. Within a sampling fraction of 5 \%, the average distance to a sampled node reaches a fixed distance. }}
\end{figure}

\section{Conclusions}
We have here shown that the graph based AL can be divided into two or three main phases. In the most early phase, when the fraction of the tagged nodes is very small (much smaller than one over the average degree), the best nodes to query are nodes that can maximize the propagation of information into yet unstudied regions. As the graphs is more densely sampled, a regional approach is optimal, where the best nodes to query are those in regions of high uncertainty (e.g. the regional entropy approach proposed here). At such intermediate tagged fractions, the main goal of querying the oracle is to find enough samples in the vicinity of each node with high uncertainty. Beyond such a fraction, only the node itself is of importance, and graph independent approaches can be used.

There are many active learning approaches including: uncertainty, representative, influence, error reduction.
It has been shown by others that usually hybrid techniques, combining several approaches, outperform using only one approach.
We here proposed two novel measures, regional uncertainty and adaptive page rank. Those two measures can be combined with existing active learning techniques to achieve even better performance.

While we have studied here only graph based ML, the same approach can be used in any task where a distance metric can be defined on the input samples. In such cases, the region of each node can be defined using distances instead of edges.

\medskip
\bibliographystyle{ieeetr}
\bibliography{AL_GCN}

\begin{thebibliography}{10}

\bibitem{ji2012variance}
M.~Ji and J.~Han, ``A variance minimization criterion to active learning on
  graphs,'' in {\em Artificial Intelligence and Statistics}, pp.~556--564,
  2012.

\bibitem{berberidis2018data}
D.~Berberidis and G.~B. Giannakis, ``Data-adaptive active sampling for
  efficient graph-cognizant classification,'' {\em IEEE Transactions on Signal
  Processing}, vol.~66, no.~19, pp.~5167--5179, 2018.

\bibitem{zhu2003semi}
X.~Zhu, Z.~Ghahramani, and J.~D. Lafferty, ``Semi-supervised learning using
  gaussian fields and harmonic functions,'' in {\em Proceedings of the 20th
  International conference on Machine learning (ICML-03)}, pp.~912--919, 2003.

\bibitem{zhu2003combining}
X.~Zhu, J.~Lafferty, and Z.~Ghahramani, ``Combining active learning and
  semi-supervised learning using gaussian fields and harmonic functions,'' in
  {\em ICML 2003 workshop on the continuum from labeled to unlabeled data in
  machine learning and data mining}, vol.~3, 2003.

\bibitem{sindhwani2005beyond}
V.~Sindhwani, P.~Niyogi, and M.~Belkin, ``Beyond the point cloud: from
  transductive to semi-supervised learning,'' in {\em Proceedings of the 22nd
  international conference on Machine learning}, pp.~824--831, ACM, 2005.

\bibitem{belkin2004semi}
M.~Belkin and P.~Niyogi, ``Semi-supervised learning on riemannian manifolds,''
  {\em Machine learning}, vol.~56, no.~1-3, pp.~209--239, 2004.

\bibitem{shi2000normalized}
J.~Shi and J.~Malik, ``Normalized cuts and image segmentation,'' {\em
  Departmental Papers (CIS)}, p.~107, 2000.

\bibitem{yang2013community}
J.~Yang, J.~McAuley, and J.~Leskovec, ``Community detection in networks with
  node attributes,'' in {\em 2013 IEEE 13th International Conference on Data
  Mining}, pp.~1151--1156, IEEE, 2013.

\bibitem{rosen2015topological}
Y.~Rosen and Y.~Louzoun, ``Topological similarity as a proxy to content
  similarity,'' {\em Journal of Complex Networks}, vol.~4, no.~1, pp.~38--60,
  2015.

\bibitem{naaman2018edge}
R.~Naaman, K.~Cohen, and Y.~Louzoun, ``Edge sign prediction based on a
  combination of network structural topology and sign propagation,'' {\em
  Journal of Complex Networks}, vol.~7, no.~1, pp.~54--66, 2018.

\bibitem{kipf2016semi}
T.~N. Kipf and M.~Welling, ``Semi-supervised classification with graph
  convolutional networks,'' {\em arXiv preprint arXiv:1609.02907}, 2016.

\bibitem{schlichtkrull2018modeling}
M.~Schlichtkrull, T.~N. Kipf, P.~Bloem, R.~Van Den~Berg, I.~Titov, and
  M.~Welling, ``Modeling relational data with graph convolutional networks,''
  in {\em European Semantic Web Conference}, pp.~593--607, Springer, 2018.

\bibitem{benami2019topological}
I.~Benami, K.~Cohen, O.~Nagar, and Y.~Louzoun, ``Topological based
  classification of paper domains using graph convolutional networks,'' {\em
  arXiv preprint arXiv:1904.07787}, 2019.

\bibitem{settles2009active}
B.~Settles, ``Active learning literature survey,'' tech. rep., University of
  Wisconsin-Madison Department of Computer Sciences, 2009.

\bibitem{lewis1994heterogeneous}
D.~D. Lewis and J.~Catlett, ``Heterogeneous uncertainty sampling for supervised
  learning,'' in {\em Machine learning proceedings 1994}, pp.~148--156,
  Elsevier, 1994.

\bibitem{culotta2005reducing}
A.~Culotta and A.~McCallum, ``Reducing labeling effort for structured
  prediction tasks,'' in {\em AAAI}, vol.~5, pp.~746--751, 2005.

\bibitem{settles2008analysis}
B.~Settles and M.~Craven, ``An analysis of active learning strategies for
  sequence labeling tasks,'' in {\em Proceedings of the conference on empirical
  methods in natural language processing}, pp.~1070--1079, Association for
  Computational Linguistics, 2008.

\bibitem{berg2017graph}
R.~v.~d. Berg, T.~N. Kipf, and M.~Welling, ``Graph convolutional matrix
  completion,'' {\em arXiv preprint arXiv:1706.02263}, 2017.

\bibitem{grover2016node2vec}
A.~Grover and J.~Leskovec, ``node2vec: Scalable feature learning for
  networks,'' in {\em Proceedings of the 22nd ACM SIGKDD international
  conference on Knowledge discovery and data mining}, pp.~855--864, ACM, 2016.

\bibitem{scheffer2001active}
T.~Scheffer, C.~Decomain, and S.~Wrobel, ``Active hidden markov models for
  information extraction,'' in {\em International Symposium on Intelligent Data
  Analysis}, pp.~309--318, Springer, 2001.

\bibitem{shannon1948mathematical}
C.~E. Shannon, ``A mathematical theory of communication,'' {\em Bell system
  technical journal}, vol.~27, no.~3, pp.~379--423, 1948.

\bibitem{fujii1998selective}
A.~Fujii, T.~Tokunaga, K.~Inui, and H.~Tanaka, ``Selective sampling for
  example-based word sense disambiguation,'' {\em Computational Linguistics},
  vol.~24, no.~4, pp.~573--597, 1998.

\bibitem{nguyen2004active}
H.~T. Nguyen and A.~Smeulders, ``Active learning using pre-clustering,'' in
  {\em Proceedings of the twenty-first international conference on Machine
  learning}, p.~79, ACM, 2004.

\bibitem{zhu2009active}
J.~Zhu, H.~Wang, B.~K. Tsou, and M.~Ma, ``Active learning with sampling by
  uncertainty and density for data annotations,'' {\em IEEE Transactions on
  audio, speech, and language processing}, vol.~18, no.~6, pp.~1323--1331,
  2009.

\bibitem{macskassy2009using}
S.~A. Macskassy, ``Using graph-based metrics with empirical risk minimization
  to speed up active learning on networked data,'' in {\em Proceedings of the
  15th ACM SIGKDD international conference on Knowledge discovery and data
  mining}, pp.~597--606, ACM, 2009.

\bibitem{ping2017batch}
S.~Ping, D.~Liu, B.~Yang, Y.~Zhu, H.~Chen, and Z.~Wang, ``Batch mode active
  learning for node classification in assortative and disassortative
  networks,'' {\em IEEE Access}, vol.~6, pp.~4750--4758, 2017.

\bibitem{cai2017active}
H.~Cai, V.~W. Zheng, and K.~C.-C. Chang, ``Active learning for graph
  embedding,'' {\em arXiv preprint arXiv:1705.05085}, 2017.

\bibitem{ma2013sigma}
Y.~Ma, R.~Garnett, and J.~Schneider, ``$\sigma$-optimality for active learning
  on gaussian random fields,'' in {\em Advances in Neural Information
  Processing Systems}, pp.~2751--2759, 2013.

\bibitem{zhou2014active}
J.~Zhou and S.~Sun, ``Active learning of gaussian processes with
  manifold-preserving graph reduction,'' {\em Neural Computing and
  Applications}, vol.~25, no.~7-8, pp.~1615--1625, 2014.

\bibitem{chen2019activehne}
X.~Chen, G.~Yu, J.~Wang, C.~Domeniconi, Z.~Li, and X.~Zhang, ``Activehne:
  Active heterogeneous network embedding,'' {\em arXiv preprint
  arXiv:1905.05659}, 2019.

\bibitem{muchnik2007self}
L.~Muchnik, R.~Itzhack, S.~Solomon, and Y.~Louzoun, ``Self-emergence of
  knowledge trees: Extraction of the wikipedia hierarchies,'' {\em Physical
  Review E}, vol.~76, no.~1, p.~016106, 2007.

\bibitem{malliaros2016locating}
F.~D. Malliaros, M.-E.~G. Rossi, and M.~Vazirgiannis, ``Locating influential
  nodes in complex networks,'' {\em Scientific reports}, vol.~6, p.~19307,
  2016.

\bibitem{page1999pagerank}
L.~Page, S.~Brin, R.~Motwani, and T.~Winograd, ``The pagerank citation ranking:
  Bringing order to the web.,'' tech. rep., Stanford InfoLab, 1999.

\bibitem{mccallum2000automating}
A.~K. McCallum, K.~Nigam, J.~Rennie, and K.~Seymore, ``Automating the
  construction of internet portals with machine learning,'' {\em Information
  Retrieval}, vol.~3, no.~2, pp.~127--163, 2000.

\bibitem{sen2008collective}
P.~Sen, G.~Namata, M.~Bilgic, L.~Getoor, B.~Galligher, and T.~Eliassi-Rad,
  ``Collective classification in network data,'' {\em AI magazine}, vol.~29,
  no.~3, pp.~93--93, 2008.

\bibitem{giles1998citeseer}
C.~L. Giles, K.~D. Bollacker, and S.~Lawrence, ``Citeseer: An automatic
  citation indexing system.,'' in {\em ACM DL}, pp.~89--98, 1998.

\bibitem{leskovec2007graph}
J.~Leskovec, J.~Kleinberg, and C.~Faloutsos, ``Graph evolution: Densification
  and shrinking diameters,'' {\em ACM Transactions on Knowledge Discovery from
  Data (TKDD)}, vol.~1, no.~1, p.~2, 2007.

\bibitem{snapnets}
J.~Leskovec and A.~Krevl, ``{SNAP Datasets}: {Stanford} large network dataset
  collection.'' \url{http://snap.stanford.edu/data}, June 2014.

\bibitem{yin2017local}
H.~Yin, A.~R. Benson, J.~Leskovec, and D.~F. Gleich, ``Local higher-order graph
  clustering,'' in {\em Proceedings of the 23rd ACM SIGKDD International
  Conference on Knowledge Discovery and Data Mining}, pp.~555--564, ACM, 2017.

\bibitem{konect:2017:subelj_cora}
``Cora citation network dataset -- {KONECT},'' Apr. 2017.

\bibitem{west2012human}
R.~West and J.~Leskovec, ``Human wayfinding in information networks,'' in {\em
  Proceedings of the 21st international conference on World Wide Web},
  pp.~619--628, ACM, 2012.

\bibitem{west2009wikispeedia}
R.~West, J.~Pineau, and D.~Precup, ``Wikispeedia: An online game for inferring
  semantic distances between concepts,'' in {\em Twenty-First International
  Joint Conference on Artificial Intelligence}, 2009.

\bibitem{mcdowell2015relational}
L.~K. McDowell, ``Relational active learning for link-based classification,''
  in {\em 2015 IEEE International Conference on Data Science and Advanced
  Analytics (DSAA)}, pp.~1--10, IEEE, 2015.

\bibitem{bilgic2010active}
M.~Bilgic, L.~Mihalkova, and L.~Getoor, ``Active learning for networked data,''
  in {\em Proceedings of the 27th international conference on machine learning
  (ICML-10)}, pp.~79--86, 2010.

\end{thebibliography}

\end{document}